\documentclass[preprint,5p,times,twocolumn]{elsarticle}
\usepackage{epsfig}

\usepackage{amssymb}
\usepackage{amsmath}
\usepackage{graphicx}
\usepackage{multirow}
\usepackage{hyperref}
\usepackage{algorithm}
\usepackage{booktabs}
\usepackage{algorithmic}
\usepackage{array}       
\usepackage{makecell} 

\journal{Biomedical Signal Processing and Control}

\begin{document}

\begin{frontmatter}
\title{SpinalSAM-R1: A Vision-Language Multimodal Interactive System for Spine CT Segmentation}
\author[label1]{Jiaming Liu}
\author[label1]{Dingwei Fan}
\author[label1]{Junyong Zhao}
\author[label2,label3]{Chunlin Li}
\author[label2,label3]{Haipeng Si\corref{cor1}}
\ead{sihaipeng1978@email.sdu.edu.cn}
\author[label1]{Liang Sun\corref{cor2}}
\ead{sunl@nuaa.edu.cn}

\affiliation[label1]{organization={College of Artificial Intelligence, Nanjing University of Aeronautics and Astronautics},%
            addressline={the Key Laboratory of Brain-Machine Intelligence Technology, Ministry of Education}, 
            city={Nanjing},
            postcode={211106}, 
            country={China}}

\affiliation[label2]{organization={Department of Orthopedics, Qilu Hospital, Shandong University},%
            addressline={}, 
            city={Jinan},
            postcode={250000}, 
            state={Shandong},
            country={China}}

\affiliation[label3]{organization={Key Laboratory of Qingdao in Medicine and Engineering},%
            addressline={Department of Orthopedics, Qilu Hospital (Qingdao), Shandong University}, 
            city={Qingdao},
            postcode={266035}, 
            state={Shandong},
            country={China}}
\cortext[cor1]{Corresponding author}
\cortext[cor2]{Corresponding author}

\begin{abstract}
The anatomical structure segmentation of the spine and adjacent structures from computed tomography (CT) images is a key step for spinal disease diagnosis and treatment. However, the segmentation of CT images is impeded by low contrast and complex vertebral boundaries. Although advanced models such as the Segment Anything Model (SAM) have shown promise in various segmentation tasks, their performance in spinal CT imaging is limited by high annotation requirements and poor domain adaptability. To address these limitations, we propose SpinalSAM-R1, a multimodal vision-language interactive system that integrates a fine-tuned SAM with DeepSeek-R1, for spine CT image segmentation. Specifically, our SpinalSAM-R1 introduces an anatomy-guided attention mechanism to improve spine segmentation performance, and a semantics-driven interaction protocol powered by DeepSeek-R1, enabling natural language-guided refinement. The SpinalSAM-R1 is fine-tuned using Low-Rank Adaptation (LoRA) for efficient adaptation. We validate our SpinalSAM-R1 on the spine anatomical structure with CT images. Experimental results suggest that our method achieves superior segmentation performance. Meanwhile, we develop a PyQt5-based interactive software, which supports point, box, and text-based prompts. The system supports 11 clinical operations with 94.3\% parsing accuracy and sub-800 ms response times. The software is released on \url{https://github.com/6jm233333/spinalsam-r1}.
\end{abstract}






\end{frontmatter}

\section{Introduction}
\label{sec:intro}

The spine's critical role in physiological function has made it a key focus in global diagnostic practices. The increasing prevalence of spinal diseases necessitates efficient imaging analysis~\cite{spinal,a01}. Computed Tomography (CT) has become the widely used imaging tool for diagnosing spinal diseases due to its high resolution and multiplanar imaging capabilities~\cite{a02,a03,a04,a05}. However, spinal CT image segmentation faces numerous challenges, including low grayscale distribution of vertebrae, complex edge morphology, mixed background tissues, and noise interference~\cite{a42}. Traditional segmentation methods are unable to handle clinical requirements~\cite{a06}. In recent years, the rapid development of artificial intelligence technologies has provided new opportunities for medical image segmentation, offering new possibilities for spinal imaging analysis, especially in the field of deep learning-driven image segmentation, where various innovative methods have emerged.

\begin{figure*}[t]
  \centering
  \includegraphics[width=\textwidth]{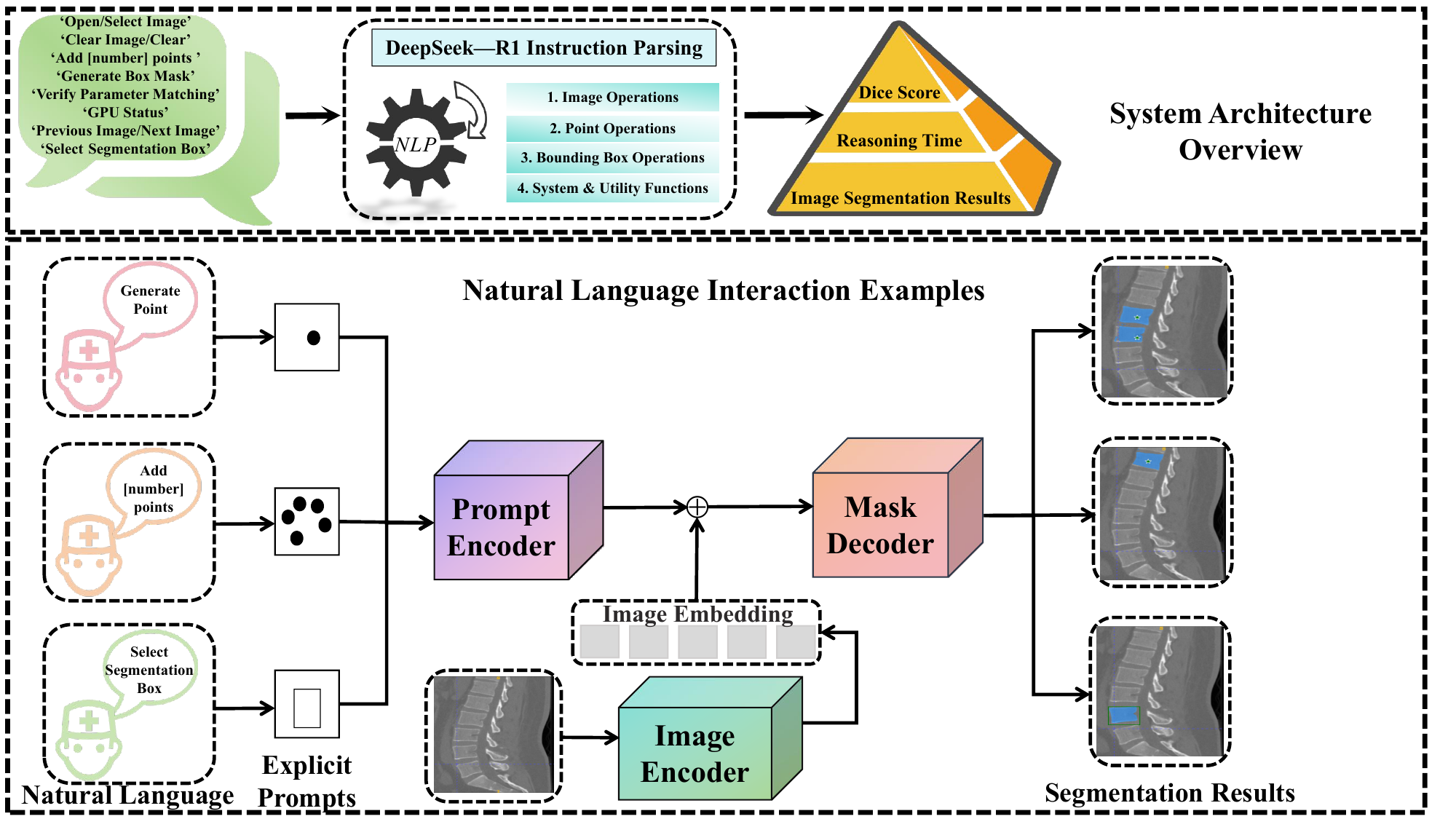} 
  \caption{Overview of the SpinalSAM-R1 system, divided into two functional blocks. \textbf{Top Block (System Architecture Overview)}: Illustrates the pipeline of instruction parsing and result evaluation—natural language commands (e.g., “Open Image”, “Add [Lumbar] points”) are processed by DeepSeek-R1 into four operation categories (Image Operations, Point Operations, etc.), then output segmentation results with metrics like Dice Score and Reasoning Time. \textbf{Bottom Block (Natural Language Interaction Examples)}: Demonstrates how explicit natural language prompts (e.g., “Generate Spine”, “Add [Lumbar] points”) are encoded via Prompt Encoder, fused with image embeddings from Image Encoder, and decoded by Mask Decoder to generate spinal segmentation results.}
  \label{flow}
\end{figure*}

The Segment Anything Model (SAM)~\cite{a09} enables interactive segmentation via multimodal prompts (points, boxes, masks), but its application in medical imaging is limited by high annotation demands and poor domain adaptability—SAM requires substantial effort to annotate high-dimensional medical images, and its accuracy for complex structures is compromised by a large natural-medical domain gap. To address the specific needs of medical imaging, researchers have made various improvements to SAM, resulting in medical-specific models. For instance, SAM-Med2D~\cite{a10} enhanced adaptability to multimodal imaging through the SA-Med2D-20M dataset, which covers 4.6 million medical images, supporting segmentation of various anatomical structures in CT, MRI, and other modalities. MA-SAM~\cite{a13} introduced a multi-atlas pseudo-prompt generation strategy, leveraging anatomical priors to improve spinal segmentation accuracy and reducing per-case processing time by 83\%.  However, these models rely primarily on visual prompts with limited natural language support. Although SAM's ViT-H backbone overfits on small medical datasets~\cite{Ma2025-SAM-Multimodal}, we address this by integrating LoRA-based fine-tuning with an anatomy-guided CBAM module, retaining feature representation while avoiding overfitting through constrained updates and anatomical guidance. Recent works such as LISA~\cite{LISA}, GROUNDHOG~\cite{GROUNDHOG}, HuggingGPT~\cite{a38}, and Visual ChatGPT~\cite{a39} have begun integrating LLMs for segmentation tasks, demonstrating the versatility of combining language models with visual foundation models to address multifaceted visual problems. 

Recent works such as LISA~\cite{LISA}, GROUNDHOG~\cite{GROUNDHOG}, HuggingGPT~\cite{a38}, and Visual ChatGPT~\cite{a39} have begun integrating LLMs for segmentation tasks, demonstrating the versatility of combining language models with visual foundation models to address multifaceted visual problems. However, despite these advances in general image segmentation, their application to medical image segmentation remains relatively underexplored~\cite{LISA,GROUNDHOG}. 


To address these challenges in spinal CT segmentation tasks, we propose a novel spine image segmentation framework that leverages the strengths of DeepSeek-R1 in conjunction with the Segment Anything model, fine-tuned specifically on a spine CT image dataset to better meet the demands of medical imaging tasks, called SpinalSAM-R1, illustrated in Fig.~\ref{flow}. Specifically, our SpinalSAM-R1 is a visual-language multimodal interactive system structured with a three-layer architecture. The user interface layer supports flexible multimodal prompts, including points, bounding boxes, and natural language commands, enabling intuitive clinical interaction.

The business logic layer integrates a fine-tuned Segment Anything Model (SAM) with the DeepSeek-R1 natural language processing module, allowing dynamic interpretation of semantic instructions into precise segmentation prompts and providing real-time, context-aware mask refinement. The infrastructure layer handles efficient data preprocessing and caching, and leverages GPU-accelerated computation to meet the demands of large-scale medical image processing with high responsiveness. The system was rigorously evaluated on a clinical dataset that comprises 120 lumbar CT scans (31,454 slices) from Shandong University Qilu Hospital. By windowing, slice filtering, and dataset splitting procedures, SpinalSAM-R1 achieved a Dice coefficient of 0.9532 and an IoU of 0.9114, surpassing state-of-the-art methods such as U-Net, TransUNet, and SAM-Med2D variants. The DeepSeek-R1 module further enhanced usability, achieving 94.3\% command parsing accuracy for 11 clinical operation types (e.g., ``Add three points'') with sub-800 ms latency.

Our major contributions to this work are summarized as follows:

\begin{enumerate}
   \item We propose an integrated SAM model enhanced with CBAM for feature refinement and LoRA for parameter-efficient fine-tuning, improving adaptability to complex spinal structures while maintaining computational efficiency.
    \item We develop a semantics-driven interaction approach by integrating the DeepSeek-R1 into medical image segmentation software. This allows users to perform segmentation tasks through natural language commands with high accuracy and low response latency, representing a significant advancement in human-computer interaction for medical applications.
    \item The constructed interaction framework ensures real-time mask rendering and cross-platform compatibility, making the system highly accessible and practical for clinical settings. The lightweight design addresses the deployment challenges associated with large models, facilitating broader application in resource-constrained environments.
\end{enumerate}

\section{Related Works}
\subsection{Deep Learning for Medical Image Segmentation}
U-Net~\cite{a07} and its variants~\cite{Horng2019} dominate medical image segmentation via multi-scale feature extraction and skip connections; UNet++~\cite{Zhou2020} further enhances feature aggregation for higher accuracy. To further enhance feature representation, attention mechanisms such as the convolutional block attention module (CBAM)~\cite{a25} have been introduced, sequentially applying channel and spatial attention to help models focus on key anatomical features and boundaries. CBAM has been successfully applied in medical image analysis~\cite{a26}. In addition, some works~\cite{a13} have explored the integration of prior knowledge and advanced regularization to further improve segmentation robustness. Despite these advances, deep learning models require large amounts of annotated data and face challenges in low-contrast medical images.

\subsection{Segment Anything in Medical Image Segmentation}
The Segment Anything Model (SAM) introduces a promptable segmentation framework built on a ViT backbone and shows broad zero-shot ability on natural images~\cite{RobustSAM2024}. However, domain gaps in texture, modality, and anatomy limit direct transfer to clinical images~\cite{Hu2025}. To bridge this, medical adaptations fine-tune SAM on large-scale medical data (e.g., MedSAM) or inject parameter-efficient modules such as LoRA to reduce trainable parameters while retaining capacity~\cite{a20}. These efforts improve accuracy and efficiency but largely remain confined to visual prompts, leaving limited support for richer interaction modalities. Nevertheless, these methods still face challenges, including limited support for diverse interaction modes and insufficient integration with natural language, motivating the exploration of multimodal solutions. Therefore, we propose SpinalSAM-R1, a multimodal interactive system that integrates the DeepSeek-R1 language model to enable natural language-guided segmentation. This combination addresses existing limitations by improving segmentation accuracy, extending interaction flexibility, and enhancing clinical applicability.

\subsection{Large Language Models for Multimodal Segmentation}
Recent advances in large language models (LLMs) have profoundly influenced natural language processing and spurred innovative methods for cross-modal interaction. Emerging systems such as SAM4MLLM~\cite{a40}, Grounded-SAM~\cite{a41}, HuggingGPT~\cite{a38}, and Visual ChatGPT~\cite{a39} demonstrate the potential of coupling LLMs with segmentation backbones for text-grounded or dialogue-driven pixel prediction~\cite{a35}. In the medical domain, several vision-language models have emerged, including MedVisionLlama~\cite{kumar2025}, GMAI-VL-R1~\cite{su2025}, LViT~\cite{Li2024}, VividMed~\cite{luo2025}, and MedCLIP-SAMv2~\cite{koleilat2025}. While these foundation models demonstrate impressive multi-task generalization, they often require substantial computational resources and may sacrifice task-specific precision. To address these limitations in spinal CT segmentation, we integrate a CBAM-enhanced SAM with LoRA-based fine-tuning and DeepSeek-R1 for natural language-guided interaction, extending beyond traditional point and box-based prompts to offer a more intuitive clinical experience while maintaining deployment efficiency for specialized clinical workflows.

\section{Methodology}
\subsection{System Overview}
\begin{figure*}[!h]
  \centering
  \includegraphics[width=0.85\textwidth]{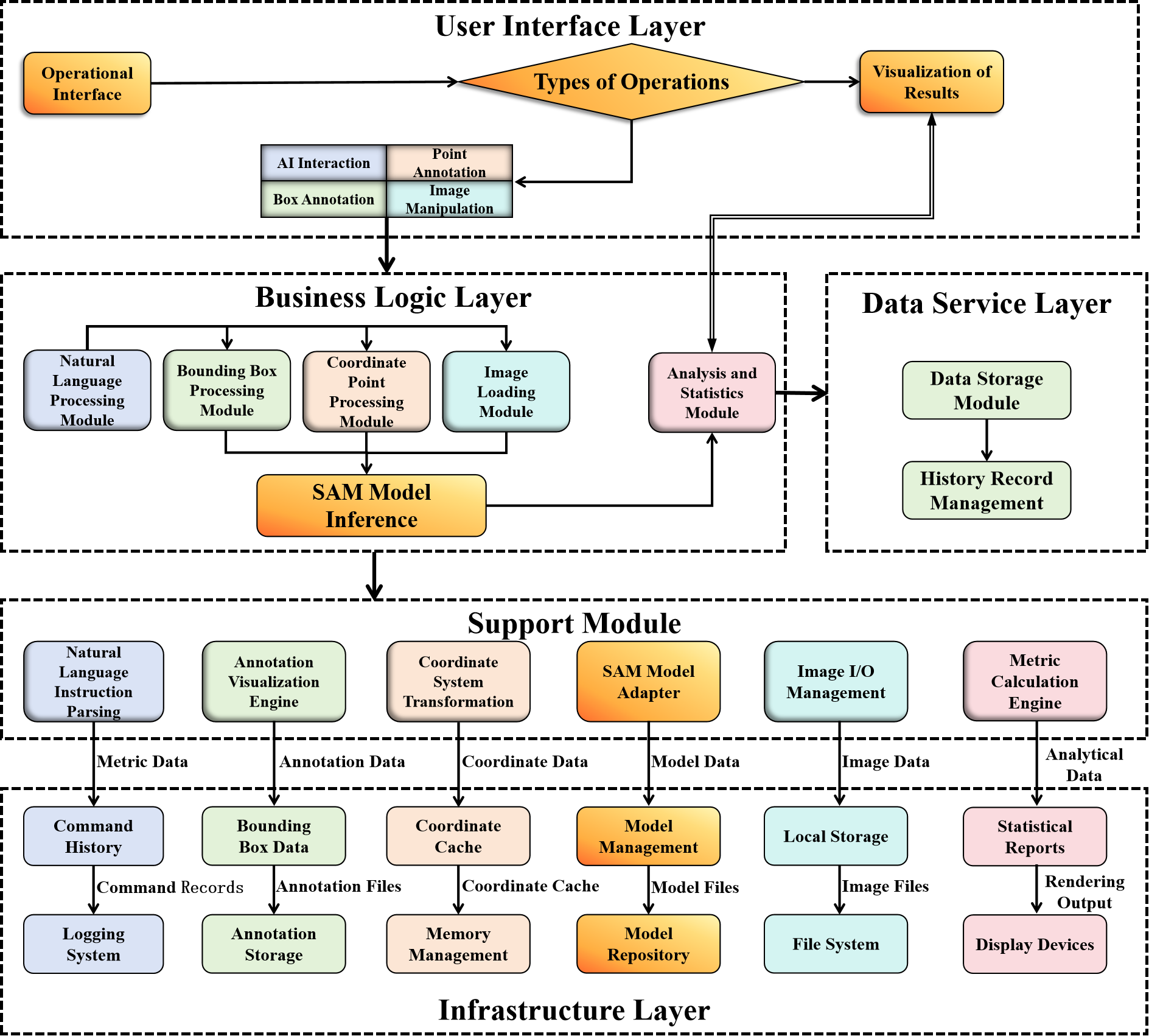} 
  \caption{System architecture of SpinalSAM-R1, comprising five hierarchical layers: \textbf{User Interface Layer} (PyQt5-based, supports point/box/text interaction with real-time visualization), \textbf{Business Logic Layer} (integrates SAM model inference and DeepSeek-R1 natural language parsing), \textbf{Data Service Layer} (manages image loading, caching, and annotation storage), \textbf{Support Module} (handles annotation visualization, coordinate transformation, and model adaptation), and \textbf{Infrastructure Layer} (governs hardware resource allocation, model deployment, and cross-platform compatibility).}
  \label{MedSAM_R1_frame}
\end{figure*}

The proposed SpinalSAM-R1 is an intelligent, multimodal segmentation system for spinal CT images, integrating a feature-enhanced SAM backbone with a large language model (DeepSeek-R1) for interactive segmentation. The overall system architecture is illustrated in Fig.~\ref{MedSAM_R1_frame}. The framework is organized into five major layers:
\begin{itemize}
    \item \textbf{User Interface Layer}: Provides a PyQt5-based interface supporting point, box, and natural language inputs with real-time visualization.
    \item \textbf{Business Logic Layer}: Integrates DeepSeek-R1 for command parsing, prompt encoding for multimodal inputs, and the enhanced SAM for segmentation inference.
    \item \textbf{Data Service Layer}: Manages image loading, caching, annotation storage, and ensures efficient data flow between interface and inference engine.
    \item \textbf{Support Module}: Handles model management, memory optimization, logging, and system monitoring for robust operation.
    \item \textbf{Infrastructure Layer}: Manages computational resources, GPU acceleration, and cross-platform deployment for high responsiveness.
\end{itemize}

\subsection{Feature-Enhanced SAM Segmentation Framework}
The core of SpinalSAM-R1 is a fine-tuned Segment Anything Model (SAM) tailored for spinal CT segmentation. Given a spine image, we use a replication operator to expand the number of channels to 3 to make it compatible with the image encoder $f_{\text{enc}}$ of SAM, i.e., the input image $I \in \mathbb{R}^{H \times W \times 3}$. The image encoder is used to extract high-level features $F$:
\begin{equation}
    F = f_{\text{enc}}(I)
\end{equation}

\subsubsection{CBAM-based Feature Enhanced}
\begin{figure*}
  \centering
  \includegraphics[width=\textwidth]{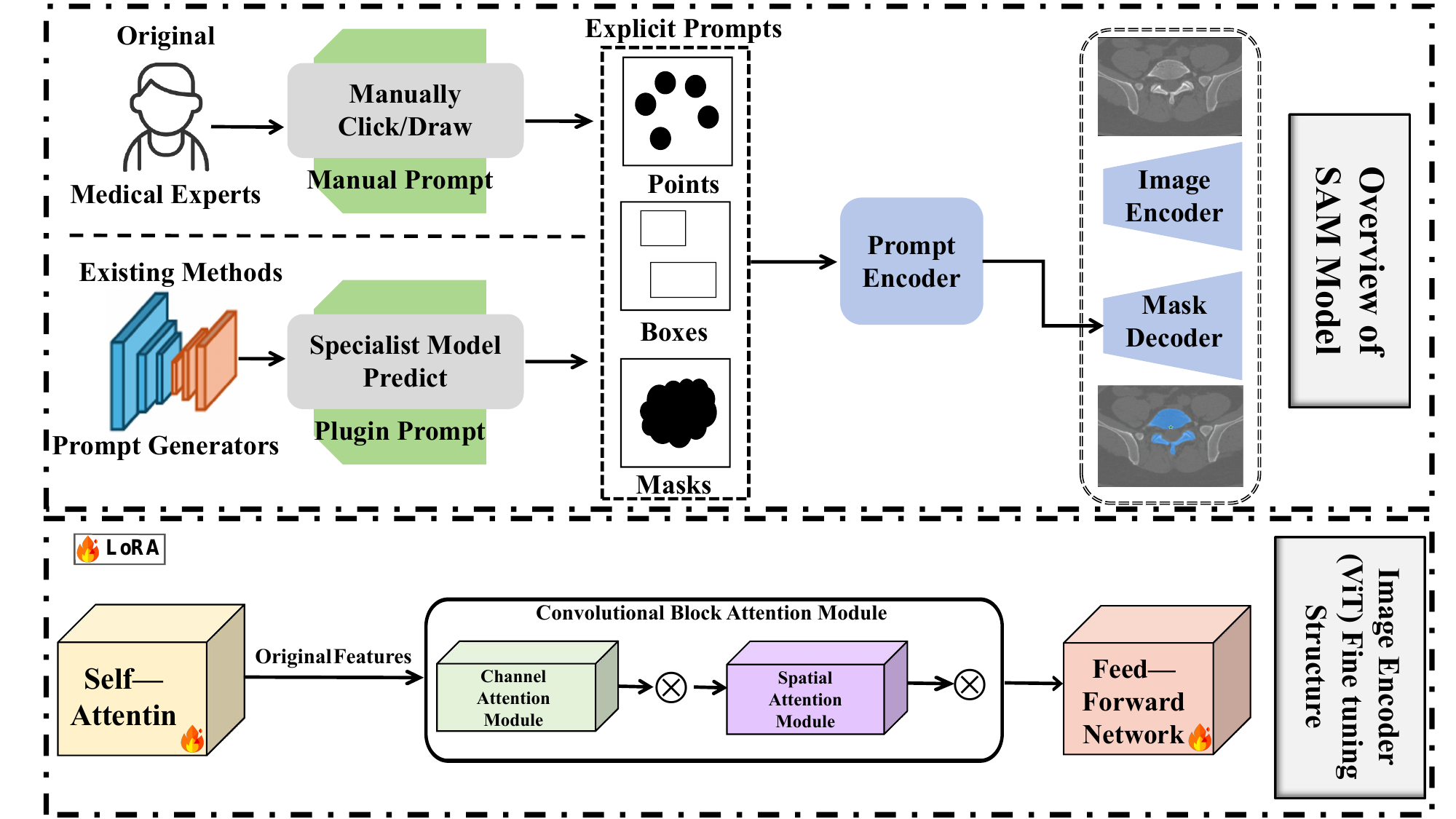} 
  \caption{Overview of feature-enhanced SAM, showing the integration of feature-enhanced SAM with multimodal user interaction.}
  \label{sam_fine}
\end{figure*}

To address the anatomical complexity and inter-subject variability in spine images, we integrate a convolutional block attention module (CBAM) into the ViT-based image encoder to enhance the anatomical structure features learning capability of the SAM's image encoder. As shown in Fig.~\ref{sam_fine}, CBAM sequentially applies channel and spatial attention to the encoder anatomical structure features, enabling the model to focus on vertebral boundaries and salient regions. The CBAM is defined as,
\begin{equation}
    F' = \text{CBAM}(F) = \text{SpatialAtt}(\text{ChannelAtt}(F))
\end{equation}
where the channel attention $\text{ChannelAtt}$ is defined as:
\begin{equation}
    \text{ChannelAtt} = \sigma \left( \text{MLP}(\text{GP}(F)) + \text{MLP}(\text{MP}(F)) \right)
\end{equation}
and the spatial attention $\text{SpatialAtt}$ is expressed as:
\begin{equation}
   \text{SpatialAtt} = \sigma \left( \text{Conv}_{7 \times 7}([\text{AP}_C(F); \text{MP}_C(F)]) \right)
\end{equation}
where $\sigma$ denotes the sigmoid function, $\text{GP}(\cdot)$ and $\text{MP}(\cdot)$ represent global average pooling and maximum pooling respectively, and $[\cdot ; \cdot]$ indicates channel-wise concatenation.

\subsubsection{LoRA-Based Fine-Tuning}
To achieve parameter-efficient adaptation to medical data, we employ Low-Rank Adaptation (LoRA) in the transformer layers to fine-tune the original SAM. The original attention weight matrix $W \in \mathbb{R}^{d \times d}$ is updated as:

\begin{equation}
    W' = W + \Delta W, \quad \Delta W = AB
\end{equation}
with $A \in \mathbb{R}^{d \times r}$ and $B \in \mathbb{R}^{r \times d}$, where $r \ll d$ is a small rank controlling additional parameters. LoRA thus enables updating only $A$ and $B$ during fine-tuning, greatly reducing the number of trainable parameters while preserving the original pre-trained weights $W$. This approach allows for effective fine-tuning with minimal additional parameters, preserving the generalization ability of the pre-trained model while adapting to the specific characteristics of spinal CT data.

The overall fine-tuning architecture, incorporating both CBAM and LoRA modules, is illustrated in Fig.~\ref{sam_fine}. The final segmentation mask $M$ is generated by fusing the refined features $F'$ with the prompt embedding $P$, where the prompt encoder $f_{\text{prompt}}$ encodes user-provided prompts (points, boxes, or text) and the mask decoder $f_{\text{dec}}$ produces the output mask:

\begin{equation}
    P = f_{\text{prompt}}(\text{prompt}), \quad M = f_{\text{dec}}(F', P)
\end{equation}

\subsubsection{Interactive Training Strategy}
We adopt an interactive training strategy to further improve segmentation performance. During the initial training round, prompts such as points or bounding boxes are randomly sampled from the ground truth to provide diverse supervision. In subsequent iterations, prompts are dynamically adjusted based on the error regions between the predicted and ground truth masks, guiding the model to focus on challenging areas. During the first round, all parameters are updated, while in later rounds, only the mask decoder is optimized, which accelerates convergence and enhances adaptability to complex segmentation tasks.
 
This feature-enhanced SAM backbone forms the foundation for accurate and robust segmentation in SpinalSAM-R1. By integrating anatomical attention and parameter-efficient adaptation, the model is well-suited for the challenges of spinal CT image segmentation. Building upon this backbone, we further develop a multimodal interactive system that enables seamless integration of natural language processing and user interaction, as described in the following section.

\subsection{Multimodal Interactive System with DeepSeek-R1 Integration}
\begin{figure}[!h]
  \centering
  \includegraphics[width=0.5\textwidth]{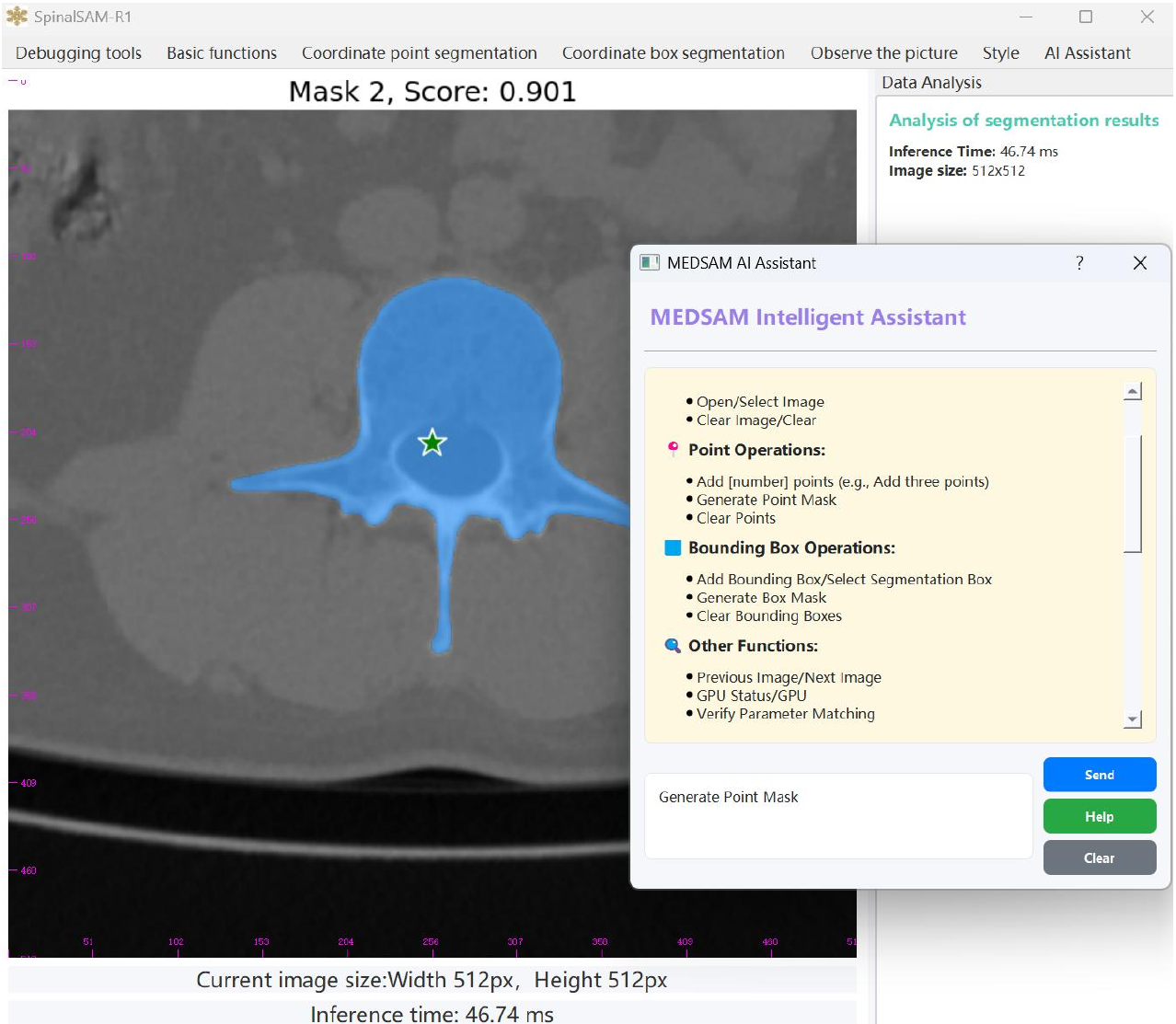} 
  \caption{The SpinalSAM-R1 multimodal system supports interactive segmentation via manual annotation or natural language commands.}
  \label{SpinalSAM_R1_Software}
\end{figure}

As shown in Fig.~\ref{SpinalSAM_R1_Software}, our SpinalSAM-R1 also provides a user-friendly and efficient method based on the large language model. Specifically, the SpinalSAM-R1 adopts a three-layer system architecture that supports multimodal interaction and seamless integration of natural language processing. The user interface layer, implemented with PyQt5, enables high-resolution image display and real-time coordinate feedback, allowing users to interact with the system through annotation tools or natural language commands. The business logic layer is responsible for managing prompt encoding, model inference, and command parsing. In particular, it leverages the DeepSeek-R1 module to parse natural language instructions into structured prompts that the segmentation model can directly utilize. The infrastructure layer handles data storage, model loading (supporting both CUDA and CPU), and hardware optimization, ensuring robust and efficient system operation.

The integration of DeepSeek-R1 and SAM enables natural language-driven segmentation in a closed-loop workflow. When a user inputs a natural language command $S$ (e.g., ``Add bounding box''), DeepSeek-R1 parses the command into a structured prompt $p$:
\begin{equation}
    p = \text{DeepSeek-R1}(S)
\end{equation}
This prompt is then encoded and fed into the SAM model, which generates candidate segmentation masks. The system automatically ranks these masks by confidence and selects the best result. Finally, the system provides the user with feedback, including both visual segmentation results and quantitative metrics, thereby facilitating intuitive and effective interaction for clinical applications.

\subsection{Loss Function}
The loss function was designed to address class imbalance and enhance boundary precision by combining Focal Loss\cite{a17} and Dice Loss\cite{a18} in a 1:1 ratio. The combined loss function is defined as follows:
\begin{equation}
    \mathcal{L}_{Total} = \mathcal{L}_{Focal} + \mathcal{L}_{Dice} 
\end{equation}
The individual components of the Focal loss and Dice loss functions are defined as:
\begin{equation}
\mathcal{L}_{Focal} = -\alpha (1 - p_t)^\gamma \phi \log(p_t)
\end{equation}
where $\phi$ is the ground truth label, $p_t$ is the predicted probability of the positive class, $\alpha$ is the balancing parameter, and $\gamma$ is the focusing parameter.
\begin{equation}
    \mathcal{L}_{Dice} = 1 - \frac{2 \times | GT\cap Pred |}{| GT | + | Pred |}
\end{equation}
where $GT$ represents the ground truth mask, and $Pred$ denotes the predicted mask. The numerator $2 \times | GT\cap Pred |$ calculates the intersection between the ground truth and predicted regions, while the denominator $| GT | + | Pred |$ represents the sum of the sizes of the ground truth and predicted regions.

\section{Materials and Methods}
\label{experiment}

\subsection{Dataset and Preprocessing}
The spine CT imaging dataset used in this study was collected from Shandong University Qilu Hospital, comprising 120 lumbar CT scans from 120 patients, including 31,454 2D slices. Each 3D scan has a voxel spacing of 0.5$\times$0.5$\times$1 mm and a resolution of 512$\times$512 pixels, with the number of slices per scan ranging from 219 to 326. The dataset is annotated for three categories: background, intervertebral disc (IVB), and vertebra (VB). Annotations were initially performed by junior experts and subsequently refined by senior experts using ITK-SNAP~\cite{Yushkevich2017,a16}. In this study, we focus solely on the binary segmentation of intervertebral discs (IVB) versus background, aiming to isolate the disc structures with high precision while ignoring the nerve and vertebral annotations.

Data preprocessing was conducted as follows. First, windowing was applied to each CT image to map a specific Hounsfield Unit (HU) range (e.g., bone window for the spine) to the normalized range [0,1]:
\begin{equation}
    I_{w} = \text{Clip}\left( \frac{I - W_{\text{Level}} + 0.5 \times W_{\text{Width}}}{W_{\text{Width}}}, 0, 1 \right)
\end{equation}
where $I$ denotes the original CT pixel values, $W_{\text{Width}}$ controls the displayed grayscale range, and $W_{\text{Level}}$ sets the window level. The $\text{Clip}$ function ensures the output is within $[0,1]$, resulting in normalized image data $I_{w}$.

Next, slices were extracted along the sagittal, coronal, and axial planes. Slices with a short side that was less than half the length of the long side were discarded to ensure anatomical consistency. Single-class masks were generated from multi-class masks, and samples with a target region area less than 1\% of the image area were excluded. The target region is calculated as:
\begin{equation}
    A_{\text{target}} = \sum_{i=1}^{H} \sum_{j=1}^{W} M(i, j)
\end{equation}
where $M(i, j)$ is the mask value at position $(i, j)$, and $H$ and $W$ are the image dimensions. Samples with $A_{\text{target}} < 0.01 \times H \times W$ were removed.

Finally, the filtered dataset was randomly split 8:2 into training/testing sets, and all images were resized to 512$\times$512 pixels for model input.

\subsection{Competing Methods}

To evaluate the effectiveness of the proposed model, we compare it with several state-of-the-art methods widely used in medical image segmentation, including U-Net~\cite{a07}, TransUNet~\cite{a08}, Swin-UNet~\cite{a19}, and SAM-Med2D. The above competing methods cover a broad spectrum from classical CNNs, hybrid CNN-Transformer architectures to pure Transformer-based and foundation model approaches. This allows a comprehensive evaluation of our proposed method against contemporary networks with varying design philosophies and abilities to capture local and global contexts in medical images.

\subsection{Computing Infrastructure}

The SpinalSAM-R1 system was implemented and tested on an NVIDIA 4090 GPU, leveraging the PyTorch 2.0 deep learning framework for model optimization. For training, the base SAM model (SAM-H) was fine-tuned due to memory constraints, employing the Adam optimizer with an initial learning rate of $10^{-4}$. The training process spanned 1500 epochs, with all images preprocessed to a standardized resolution of $512 \times 512$ pixels during the data preparation phase. 

\subsection{Evaluation Metrics}
In our experiments, we used the Dice Coefficient (DC), Intersection over Union (IoU), Mean Surface Distance (MSD), and 95\% Hausdorff Distance (HD95) to measure the segmentation performance of all competing methods. They are described below.

\begin{equation}
   DC = \frac{2 \times | GT \cap Pred |}{| GT | + | Pred |}
\end{equation}
\begin{equation}
    IoU = \frac{| GT \cap Pred |}{| GT \cup Pred |}
\end{equation}

\begin{equation}
MSD = \frac{\sum_{p \in \text{GT}} \min_{q \in \text{Pred}} \| p, q \| + \sum_{q \in \text{Pred}} \min_{p \in \text{GT}} \| p, q \|}{N_{\text{GT}} + N_{\text{Pred}}}
\end{equation}
\begin{equation}
HD95 = \text{Max} \left( \sup_{p \in \text{GT}} \inf_{q \in \text{Pred}} \| p, q \|, \sup_{q \in \text{Pred}} \inf_{p \in \text{GT}} \| p ,q \| \right)
\end{equation}
where GT represents the ground truth of the input image, and Pred represents the predicted segmentation result. $N_{\text{GT}}$ and $N_{\text{Pred}}$ represent the number of points on the ground truth surface $\text{GT}$ and the predicted surface $\text{Pred}$, respectively. $p$ and $q$ denote individual points on the ground truth surface and the predicted surface, respectively, and $\| \cdot \|$ represents the Euclidean distance.

\section{Results and Discussion}
\label{results}

\subsection{Experimental Results}

We first evaluate our SpinalSAM-R1 for spine CT image segmentation. The results achieved by SpinalSAM-R1 and its competing methods are reported in Table~\ref{MedSAM_table}.

\begin{table*}[t]
  \centering
  \caption{Comparison of segmentation results between our method and other baseline methods on the spine dataset. Paired t-tests were conducted to assess statistical significance. The symbol "$\ast$" indicates that our method achieved statistically significant improvements ($p<0.05$). The symbol "$\uparrow$" denotes that a higher value indicates better performance, while "$\downarrow$" indicates the opposite.}
    \begin{tabular}{ccccc}
    \toprule
   Methods &  DC$\uparrow$ & IoU$\uparrow$ & MSD$\downarrow$ & HD95$\downarrow$ \\
   \midrule
   U-Net   & 0.8700 $\pm$0.0144* & 0.7861$\pm$0.0238* & 3.25$\pm$1.43* & 23.05$\pm$12.05* \\
   TransUNet  & 0.9335$\pm$0.0002*  & 0.9113$\pm$0.0005*   & 1.92$\pm$0.06*   & 5.58$\pm$2.01* \\
   Swin-UNet  & 0.8863$\pm$0.0016* & 0.9097$\pm$0.0012*   & 3.64$\pm$1.37*   & 4.79$\pm$0.02*  \\
   SAM-Med2D(Box) & 0.9316$\pm$0.0012* & 0.8738$\pm$0.0031*  & 2.25$\pm$0.54*  & 6.14$\pm$1.41*   \\
   SAM-Med2D(Point) & 0.9329$\pm$0.0011* & 0.8760$\pm$0.0029*  & 2.21$\pm$0.53*   & 6.08$\pm$4.95*  \\
   SpinalSAM-R1 (Ours)     & \textbf{0.9532$\pm$0.0005 } & \textbf{0.9114$\pm$0.0015} &\textbf{1.81}$\pm$\textbf{0.50}   & \textbf{5.47$\pm$0.73}  \\     
    \bottomrule
    \end{tabular}%
  \label{MedSAM_table}%
\end{table*}%

As shown in Table~\ref{MedSAM_table}, SpinalSAM-R1 achieves substantial improvements across all metrics (DC: 0.9532, IoU: 0.9114, MSD: 1.81, HD95: 5.47), outperforming competing methods with statistical significance ($p < 0.05$, paired t-test).

\begin{figure*}[ht]
\centering
\includegraphics[width=\textwidth]{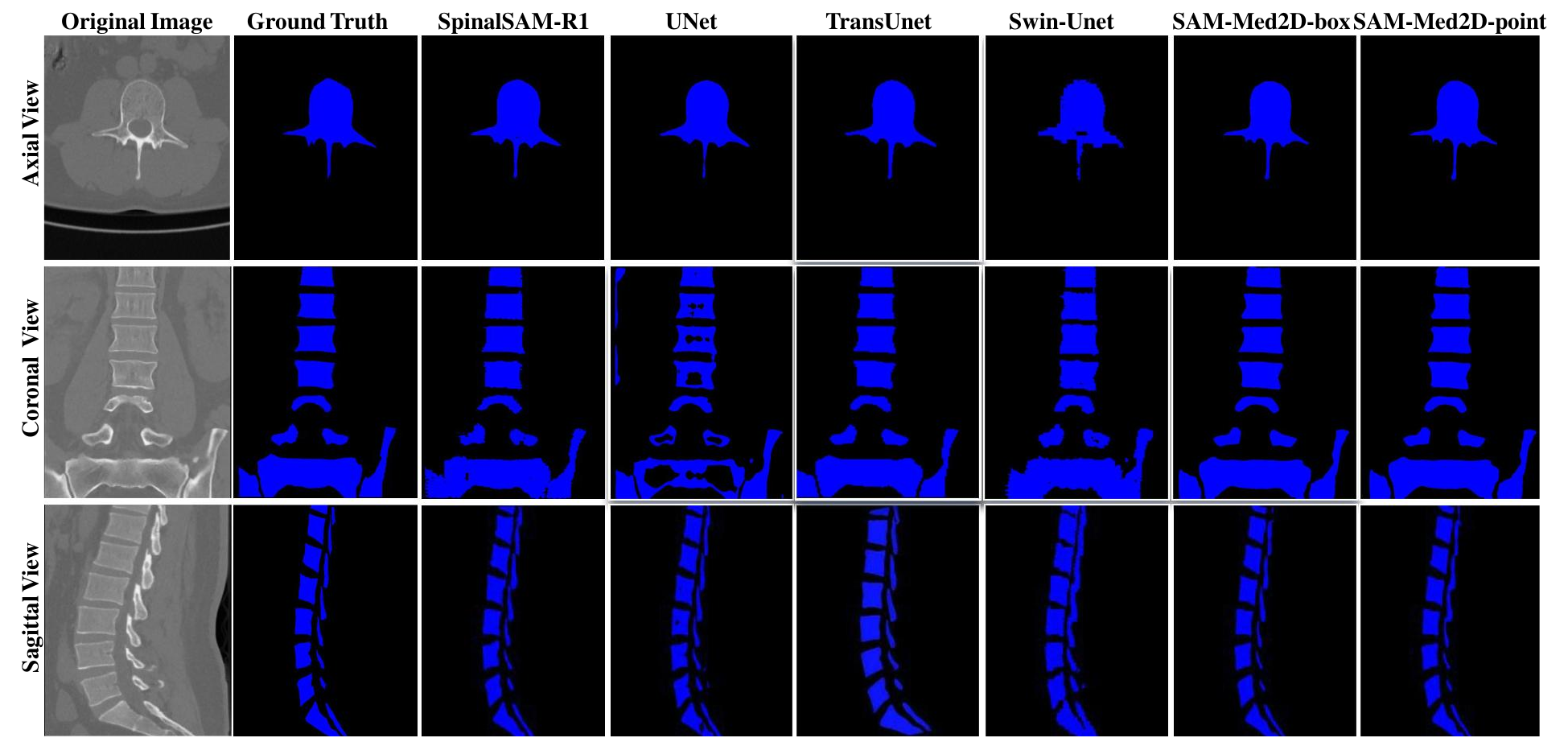} 
\caption{Segmentation results of CT lumbar images on sagittal, coronal, and axial views across different methods. From left to right, the columns show the original image, ground truth, SpinalSAM-R1, UNet, TransUNet, Swin-Unet, SAM-Med2D-box, and SAM-Med2D-point. All methods are evaluated under identical interaction prompts, with blue masks representing the predicted vertebral regions.}
\label{MedSAM_R1_Results}
\end{figure*}

Fig.~\ref{MedSAM_R1_Results} presents segmentation results across axial, coronal, and sagittal views. SpinalSAM-R1 consistently produces anatomically accurate masks across all planes, accurately capturing vertebral boundaries and fine structures. Compared to UNet, TransUNet, and Swin-Unet, which exhibit boundary discontinuities, and SAM-Med2D variants that miss subtle details, our method demonstrates superior spatial coherence and anatomical precision. In the coronal view, SpinalSAM-R1 retains strong vertebral alignment and separation, avoiding inter-vertebral leakage seen in other models—particularly the UNet and Swin-Unet, where gaps and bleed-through are notable.Box-guided SAM-Med2D reduces this issue but shows minor mask spillover, while point-based prompts lead to visible under-segmentation of lower vertebrae. Lastly, in the sagittal plane, SpinalSAM-R1 excels at maintaining structural continuity along the spine, with clearly isolated vertebrae. Competing models frequently demonstrate fused or misaligned masks, undermining anatomical fidelity. Collectively, these visualizations highlight SpinalSAM-R1's robustness in producing accurate and consistent spinal segmentation across varied perspectives.

\subsection{Ablation Study}

\begin{table*}[!h]
  \centering
  \caption{Ablation study of SpinalSAM-R1 on the spine dataset. Each row shows the effect of incrementally adding CBAM, LoRA, and interactive training.}
    \begin{tabular}{ccccccc}
    \toprule
    CBAM & LoRA &  Interactive  & DC & IoU & MSD & HD \\
    \midrule
     &  &  & 0.8850$\pm$0.0300  & 0.8000$\pm$0.0450   & 2.10$\pm$0.60   & 6.20$\pm$2.10    \\
    \checkmark  &  &  & 0.8955$\pm$0.0280  & 0.8150$\pm$0.0430   & 1.95$\pm$0.58   & 5.80$\pm$2.00 \\
    &\checkmark  &  & 0.9170$\pm$0.0270  & 0.8650$\pm$0.0410   & 1.91$\pm$0.51   & 5.52$\pm$1.85    \\
    \checkmark &\checkmark  & \checkmark & \textbf{0.9532$\pm$0.0005} & \textbf{0.9114$\pm$0.0015} & \textbf{1.81$\pm$0.50} & \textbf{5.47$\pm$0.73} \\
    \bottomrule
    \end{tabular}%
  \label{tab:ablation}%
\end{table*}%

To evaluate the impact of each module in our proposed SpinalSAM-R1, we conducted ablation experiments. Specifically, we compared our method with the original SAM, SAM with CBAM, SAM with LoRA, and our SpinalSAM-R1 (i.e., the SAM with CBAM, SAM, and Interactive). The experimental results for spine CT image segmentation are summarized in Table~\ref{tab:ablation}. 

The results show that both CBAM and LoRA independently improve segmentation performance over the baseline SAM model. Specifically, the addition of CBAM enhances the ability of SpinalSAM-R1 to focus on relevant anatomical features, which is especially beneficial for distinguishing vertebral boundaries in low-contrast regions. LoRA enables efficient fine-tuning with minimal parameter overhead, which is crucial for adapting large-scale models to limited medical datasets while maintaining generalization. When combined, these modules yield further improvements, demonstrating their complementarity. The introduction of the interactive training strategy leads to a substantial boost in all metrics, highlighting its effectiveness in improving model adaptability and robustness to diverse input prompts. These results confirm the synergy of CBAM, LoRA, and interactive training.  Integrating DeepSeek-R1 enables natural-language-driven segmentation, improving clinical usability and reducing manual annotation needs.

\subsection{Analysis of Model Parameters and Inference Time}
This software employs a deep learning-based approach for image segmentation, with the core parameters derived from the employed SAM model, specifically based on the Vit-H architecture. The total number of parameters in this model is approximately 140 million, classifying it as a large-scale pre-trained network. This considerable parameter count endows the model with robust expressive capacity, enabling it to adapt to a wide range of image segmentation tasks.

Regarding deployment and integration, the model is optimized using ONNX Runtime to facilitate high-efficiency inference. The system supports multi-threaded task scheduling and hardware acceleration, which collectively enable rapid inference on GPU-supported hardware. Empirical results demonstrate an average inference latency of approximately 250 to 300 milliseconds, ensuring the system's capability to deliver real-time responses in interactive applications.

\subsection{Core Features and User Interaction}
A fundamental aspect of our proposed system is its ability to be operated through natural language commands. Clinicians can directly input instructions such as "Open lumbar CT images", "Add three points to the vertebral body" or "Generate segmentation mask" and the system, with the help of the DeepSeek-R1 model, accurately interprets these commands and seamlessly executes the corresponding operations. Testing demonstrates that the system's recognition accuracy for relevant medical terminology, combined with an average response time within 800 milliseconds, enhances workflow efficiency and user experience in clinical scenarios.

\section{Conclusions}
\label{Conclusions}

We propose SpinalSAM-R1, an innovative vision-language multimodal system that revolutionizes spine CT segmentation through synergistic integration of enhanced medical image analysis and natural language interaction. By combining an anatomy-guided SAM architecture with DeepSeek-R1's linguistic intelligence, our system achieves state-of-the-art performance while enabling intuitive clinician interaction. Three key innovations drive this advancement: 1) A CBAM and LoRA-enhanced SAM model that maintains 99.5\% original parameters while improving vertebral segmentation; 2) The first clinical integration of an LLM-powered natural language interface supporting 11 operational commands with 94.3\% parsing accuracy; 3) An end-to-end processing framework delivering real-time segmentation (800ms latency) within a clinician-friendly interface. By bridging AI capabilities with clinical workflows, SpinalSAM-R1 establishes a new paradigm for intelligent medical imaging.

\section*{Acknowledgements}

Funding: This work was supported by the National Natural Science Foundation of China (Grants 62136004, 62376123, 62472228), National Key R\&D Program of China (Grant 2023YFF1204803), and Guangdong Basic and Applied Basic Research Foundation (Grant 2024A1515011925).

Competing interests: The authors declare that they have no conflict of interest. 

Statements of ethical approval: For this review article, the authors did not undertake work that involved human participants or animals.

\end{document}